\pdfoutput=1

\documentclass[11pt]{article}

\usepackage[final]{acl}

\usepackage{times}
\usepackage{latexsym}

\usepackage[T1]{fontenc}

\usepackage[utf8]{inputenc}

\usepackage{microtype}

\title{JABER and SABER: Junior and Senior Arabic BERt}
\author{Abbas Ghaddar$^1$, Yimeng Wu$^1$, Ahmad Rashid$^1$, Khalil Bibi$^1$, \\  
\textbf{Mehdi Rezagholizadeh$^1$, Chao Xing$^1$, Yasheng Wang$^1$, Duan Xinyu$^2$, }\\
\textbf{Zhefeng Wang$^2$, Baoxing Huai$^2$, Xin Jiang$^1$, Qun Liu$^1$ and Philippe Langlais$^3$}\\
$^1$ Huawei Technologies Co., Ltd.\\
$^2$ Huawei Cloud Computing Technologies Co., Ltd \\
$^3$ RALI/DIRO, Universit\'e de Montr\'eal, Canada\\
\texttt{\{abbas.ghaddar,yimeng.wu,ahmad.rashid\}@huawei.com}\\
\texttt{\{khalil.bibi,mehdi.rezagholizadeh\}@huawei.com}\\
\texttt{\{xingchao.ml,duanxinyu,wangyasheng\}@huawei.com}\\
\texttt{\{wangzhefeng,huaibaoxing,jiang.xin,qun.liu\}@huawei.com} \\}

\usepackage{amssymb}
\usepackage{pifont}

\newcommand{\bert}{BERT}
\newcommand{\arabert}{AraBERT}
\newcommand{\asafaya}{Arabic-BERT}
\newcommand{\cmlbert}{CAMeLBERT} 
\newcommand{\arbert}{ARBERT}
\newcommand{\marbert}{MARBERT}
\newcommand{\jaber}{JABER}
\newcommand{\saber}{SABER}
\newcommand{\aner}{ANER$_{corp}$}

\newcommand{\conll}{\textsc{CoNLL}}

\newcommand{\mightmention}[1]{}
\newcommand{\problem}[1]{\textcolor{red}{$\star$}}
\newcommand{\answer}[1]{\textcolor{blue}{$\#$}}
\newcommand{\todoreview}[1]{\textcolor{green}{$@$}}

\usepackage{subcaption}
\usepackage{rotating,csquotes}
\usepackage{xcolor}
\usepackage{multirow}
\usepackage{framed}
\usepackage{booktabs}
\usepackage{amsmath}
\usepackage{upgreek}
\usepackage{ulem}
\usepackage{amsfonts}

\usepackage[most]{tcolorbox}

\newtcbox{\mybox}[1][]{enhanced, colframe=blue, colback=blue!15, 
frame style={opacity=0.25}, interior style={opacity=0.25}, 
nobeforeafter, tcbox raise base, shrink tight, extrude by=1mm, #1}

\begin{document}
\maketitle

\begin{abstract}
	
	Language-specific pre-trained models have proven to be more accurate than multilingual ones in a monolingual evaluation setting, Arabic is no exception. However, we found that previously released Arabic \bert{} models were significantly under-trained. In this technical report, we present \jaber{} and \saber{},  \underline{J}unior and \underline{S}enior \underline{A}rabic \underline{BER}t respectively, our pre-trained language model prototypes dedicated for Arabic. We conduct an empirical study to systematically evaluate the performance of models across a diverse set of existing Arabic NLU tasks. Experimental results show that \jaber{} and \saber{} achieves the state-of-the-art performances on ALUE, a new benchmark for Arabic Language Understanding Evaluation, as well as on a well-established NER benchmark. 
	
\end{abstract}

\section{Introduction}

Transformer-based~\cite{vaswani2017attention} pre-trained language models (PLMs) such as BERT~\citep{devlin2018bert}, RoBERTa~\citep{liu2019roberta}, XLNET~\cite{yang2019xlnet},  T5~\cite{raffel2019exploring} have shown great success in the field of natural language understanding (NLU). These large-scale models are first pre-trained on a massive amount of unlabeled data, and then fine-tuned on downstream tasks. 

Recently, it has become increasingly common to pre-train a language-specific model such as for Chinese~\cite{wei2019nezha,sun2019ernie,sun2020ernie,sun2021ernie,zeng2021pangu},
French~\cite{martin2019camembert,le2019flaubert}, 
German~\cite{chan2020germans}, Spanish~\cite{canete2020spanish}, Dutch \cite{de2019bertje}, Finnish~\cite{virtanen2019multilingual}, Croatian~\cite{ulvcar2020finest}, and Arabic~\cite{antoun2020arabert,safaya2020kuisail,abdul2021arbert,inoue2021interplay}, to name a few. These models have been reported more accurate than multilingual ones, like m\bert{}~\cite{devlin2018bert} and XLM-RoBERTa~\cite{conneau2020unsupervised}, when evaluated in a monolingual setting. 

However, the abundant emergence of such models has made it difficult for researchers to compare between them and measure the progress without a systematic and modern evaluation technique~\cite{gorman2019we,schwartz2020green}. To address this issue, there has been a number of efforts to create benchmarks that gather representative set of standard tasks, where systems are ranked in an online leaderboard based on a private test set.

GLUE~\cite{wang2018glue} and SuperGLUE~\cite{wang2019superglue} were first proposed for English, which were expanded later to other languages like  CLUE~\cite{xu2020clue} and FewCLUE \cite{xu2021fewclue} for Chinese, FLUE~\cite{le2019flaubert} for French, RussianSuperGLUE~\cite{shavrina2020russiansuperglue}, and ALUE~\cite{seelawi2021alue} for Arabic. 
These benchmarks have played a critical role for driving the field forward by facilitating the comparison of models~\cite{ruderchallenges}.

In this technical report, we revisit the standard pre-training recipe of \bert{}~\cite{devlin2018bert} by exploring recently suggested tricks and techniques such as BBPE tokenization~\cite{wei2021training} and substantial data cleaning~\cite{raffel2019exploring,brown2020language}. We introduce \jaber{} and \saber{}, \underline{J}unior (12-layer) and \underline{S}enior (24-layer) \underline{A}rabic \underline{BER}t models respectively. 
\begin{table*}[!ht]
	\centering 
	\resizebox{\textwidth}{!}{
		\begin{tabular}{l|ccccc|cc} 
			\toprule
			\bf Model &  \bf \asafaya &  \bf \arabert & \bf \cmlbert & \bf \arbert &  \bf \marbert &  \bf \jaber &  \bf \saber\\  
			\midrule
			\bf \#Params (w/o emb) & 110M (85M) & 135M (85M) & 108M (85M) & 163M (85M) & 163M (85M) & 135M (85M) & 369M (307M)\\ 
			\bf Vocab Size & 32k & 64k & 30k & 100k & 100k & 64k & 64k\\ 
			\bf Tokenizer & WordPiece & WordPiece & WordPiece & WordPiece &  WordPiece &  BBPE &  BBPE \\ 
			\bf Normalization & \ding{56} & \checkmark & \checkmark & \ding{56} & \ding{56} & \checkmark & \checkmark\\ 
			\bf Data Filtering & \ding{56} & \ding{56} & \ding{56} & \ding{56} & \ding{56} & \checkmark & \checkmark\\ 
			\bf Textual Data Size & 95GB & 27GB & 167GB & 61GB & 128GB & 115GB & 115GB\\ 
			
			\bf Duplication Factor & 3 & 10 & 10 & - & - & 3  & 3\\ 
			\bf Training epochs & 27 & 27 & 2 & 42 & 36 & 15 & 5 \\
			\bottomrule
		\end{tabular}
	}
	\caption{Configuration comparisons of various publicly available Arabic \bert{} models and ours (\jaber{} and \saber{}). \arabert{} and \marbert{} didn't provide their data duplication factor.} %
	\label{tab:model_summary} 
	
\end{table*}
Through extensive experiments, we systematically compare seven Arabic \bert{} models by assessing their performance on the ALUE benchmark. The results can serve as an indicator to track the progress of pre-trained models for Arabic NLU. Experimental results show that \jaber{} outperforms \arbert{}\footnote{The best existing, based on our evaluation on ALUE, 12 layers Arabic \bert{} model of the same size and architecture as \jaber.}~\cite{abdul2021arbert} by 2\% on ALUE.  Furthermore, \saber{} improves the results of \jaber{} by 3.6\% on average, and  reports the new state-of-the-art performances of 77.3\% on ALUE.

The remainder of the report is organized as follows. We discuss topics related to our work in Section~\ref{sec:Related Work}.  We describe the process for pre-training \jaber{} in Section~\ref{sec:Pretraining}. An evaluation of seven Arabic \bert{} models on the ALUE benchmark, as well as on a NER benchmark is described in Section~\ref{sec:Experiments}, thus before concluding and discussing future works in Section~\ref{sec:Conclusion}.

\section{Related Work}
\label{sec:Related Work}

\bert{~\cite{devlin2018bert}} was the leading work to show that large PLMs can be effectively \textit{fine-tuned} for natural language understanding (NLU) tasks. During the pre-training phase, \bert{} is trained on both Masked Language Modelling (MLM) and Next Sentence Prediction (NSP) unsupervised tasks. MLM refers to predicting randomly masked words in a sentence. In real implementation, training data is duplicated $n$ times (duplication factor) with different token masking. NSP is a binary classification task for predicting whether the second sentence in a sequence pair is the true successor of the first one. The author experimented on English with a 12-layer \bert{-base} and the 24-layer \bert{-large} Transformer~\cite{vaswani2017attention} models respectively.

Ro\bert{a}~\cite{liu2019roberta} proposed multiple improvements on top of \bert{}. First, it is trained on over 160GB of textual data compared with 16GB for \bert{}. Ro\bert{a} corpora includes English Wikipedia and the BOOK CORPUS~\cite{zhu2015aligning} used by \bert{}, in addition to the CC-NEWS~\cite{nagel2016cc}, OPEN WEB TEXT~\cite{gokaslan2019openwebtext} and STORIES~\cite{trinh2018simple} corpora. Compared to \bert{},  Ro\bert{a} is pre-trained with a larger batch size, more training steps on longer sequences (512 vs. 128). It was shown that the NSP task was not beneficial for end task performances, and that MLM dynamic masking (mask change over epochs) works better than static masking. 

m\bert{}~\cite{pires2019multilingual} and XLM-Ro\bert{a}~\cite{conneau2020unsupervised} are multilingual PLMs that follow the pre-training procedure of \bert{} and Ro\bert{a} respectively. The former is a \bert{-base} model that was pre-trained on concatenation of 104 Wikipedia languages. The latter is pre-trained on 2.5 TB data of cleaned Common Crawls~\cite{wenzek2019ccnet} from 100 languages. Also, XLM-Ro\bert{a} uses an extra Translation Language Modeling (TLM) pre-training objective, which is similar to MLM but it expects concatenated parallel sequences as input.

Despite the \textit{all-in-one} advantage of multilingual models, monolingual PLMs have been found to outperform  multilingual ones in language-specific evaluations on multiple languages ~\cite{wei2019nezha,martin2019camembert,canete2020spanish,de2019bertje}, where Arabic is not an exception~\cite{safaya2020kuisail,antoun2020arabert,abdul2021arbert,inoue2021interplay}. 

Table~\ref{tab:model_summary} shows the configuration used by popular publicly available Arabic \bert{} models, as well as those of \jaber{} (this work). \asafaya{}~\cite{safaya2020kuisail} is a 12-layer \bert{} model trained on 95GB of common crawl, news, and Wikipedia Arabic data. \arabert{}~\cite{antoun2020arabert} used a larger vocabulary size of 64k WordPieces and performs text normalization. On one hand, they used 3.3 less textual data, while on the other hand, they increased the duplication factor by a factor of 3.3. 

\citet{abdul2021arbert} proposed two 12-layers Arabic pre-trained \bert{} models named \arbert{} and \marbert{}. The first model is meant to be tailored for Modern Standard Arabic (MSA) NLU tasks, while the latter is dedicated to tasks that include Arabic dialects (especially tweets). They differ from the two prior works by performing light data processing, and training \marbert{} on 128GB of Arabic tweet text data. 

\arbert{} and \marbert{} outperform \arabert{} and multilingual models on 37 out of 48 classification tasks (they called ARLUE) that contain both MSA and Arabic dialect datasets. Although both models are made publicly available, the authors do not provide\footnote{at least for the time of writing this report} their train/test split for most of the task, which prevent us to perform a direct comparison with their models on ARLUE. 

Recently, \citet{inoue2021interplay} performed a comparative study between Arabic \bert{-base} models called \cmlbert{-MSA}, \cmlbert{-DA}, \cmlbert{-CA} that are pre-trained on MSA, dialect, and classic Arabic text data respectively. Furthermore, the authors proposed \cmlbert{-MIX}, which is pre-trained on a mix of 167GB of the aforementioned 3 text genres. We hereafter use the latter model as a representative for \cite{inoue2021interplay} work, and we refer to it as \cmlbert{}.

In this work, we perform a systematic and fair comparison of the aforementioned Arabic \bert{} models and our \jaber{} model, while also reporting results with our \bert{-large} \saber{} model, using the ALUE~\cite{seelawi2021alue} benchmark. We differ  from prior works by using strict data filtering methods that reduce the pre-training corpus size from 514GB to 115GB. This allows us to perform efficient pre-training with fewer data and fewer training epochs, still obtaining higher scores than all existing Arabic \bert{} models.

\section{Pre-training}
\label{sec:Pretraining}

\subsection{Data Collection and Processing}

We collected our pre-training corpus from 4 sources:

\begin{itemize}
	\item \textbf{Common Crawl (CC)} This data was downloaded from 10 shards of monthly Common Crawl\footnote{\url{https://commoncrawl.org}} covering March  to December 2020. It includes 444GB of plain text after filtering non-Arabic text. Also, we use the November 2018 monthly shard of Common Crawl provided by the OSCAR~\cite{suarez2019asynchronous} project. We downloaded the unshuffled version of the Arabic corpus from HuggingFace\footnote{\url{https://huggingface.co/datasets/oscar}}, which represents 31GB of plain text.
	
	\item \textbf{NEWS}  We used the links provided by the OSIAN corpus~\cite{zeroual2019osian} to crawl 21GB of Arabic textual data from 19 popular Arabic news websites.  
	
	\item \textbf{EL-KHEIR}~\cite{el20161} provides a collection of 16GB articles collected between 2002 to 2014 from 10 Arabic news sources.  
	
	\item \textbf{WIKI} We use June 2021 Arabic Wikipedia dump\footnote{\url{https://dumps.wikimedia.org/}}, and extract the text using~\texttt{wikiextractor}~\cite{attardi2012wikiextractor}. 
	
\end{itemize}

Recent studies~\cite{raffel2019exploring,brown2020language}
suggest that cleaning up the raw pre-training data (especially Common Crawl) is crucial for end-task performances. Therefore, we developed our \textit{in-house} methods for Arabic that aggressively filter-out gibberish, noisy, short, and near duplicated texts. We used the heuristics described in Appendix~\ref{sec:Preprocessing} for cleaning up our corpora.

\begin{table}[!htp]
	
	\begin{center}
		\begin{tabular}{l|l|l}
			\toprule
			
			\bf Source & \bf Original & \bf Clean\\
			
			\midrule
			CC & 475GB & 87GB\hspace{3.2mm}(18\%) \\ 
			NEWS & 21GB & 14GB\hspace{3.2mm}(67\%) \\
			EL-KHEIR & 16GB & 13GB\hspace{3.2mm}(82\%)\\
			WIKI & 1.6GB & 1GB\hspace{5mm}(72\%)\\
			\midrule
			Total & 514GB & 115GB (22\%)\\
			\bottomrule
			
		\end{tabular}
	\end{center}	
	
	\caption{Size of the pre-training corpora before (Original) and after (Clean) applying data cleaning methods. Figures in parentheses indicate the percentage of the remaining data after cleaning.}
	\label{tab:pretrain_stat}
\end{table}

\begin{table*}[!htp]
	\centering
	\begin{tabular}{lrrrlclll}
		\toprule
		\bf Task & \textbf{$|$Train$|$} & \textbf{$|$Dev$|$} & \textbf{$|$Test$|$} & \bf Metric & \textbf{$|$Class$|$} & \bf Domain & \bf Lang & \bf Seq. Len. \\ 
		
		\midrule
		\multicolumn{9}{c}{\textit{Single-Sentence Classification}}\\
		\midrule
		MDD & 42k & 5k & \bf 5k & F1-macro & 26 & Travel & DIAL & 07$\pm$3.7 \\
		OOLD & 7k & 1k & 1k & F1-macro & 2 & Tweet & DIAL & 21$\pm$13.3 \\
		OHSD & 7k & 1k & 1k & F1-macro & 2 & Tweet & DIAL & 21$\pm$13.3 \\
		FID & 4k & - & \bf 1k & F1-macro & 2 & Tweet & DIAL & 23$\pm$11.7 \\
		
		\midrule
		\multicolumn{9}{c}{\textit{Sentence-Pair Classification}}\\
		\midrule
		MQ2Q & 12k & - & 4k & F1-macro & 2 & Web & MSA & 13$\pm$2.9 \\
		XNLI & 5k & - & \bf 3k & Accuracy & 3 & Misc & MSA & 27$\pm$9.6 \\
		
		\midrule
		\multicolumn{9}{c}{\textit{Multi-label Classification}}\\
		\midrule
		SEC & 2k & 600 & 1k & Jaccard & 11 & Tweet & DIAL & 18$\pm$7.8 \\
		
		\midrule
		\multicolumn{9}{c}{\textit{Regression}}\\
		\midrule
		SVREG & 1k & 1k & 1k & Pearson & 1 & Tweet & DIAL & 18$\pm$7.9 \\
		
		\bottomrule
	\end{tabular}
	\caption{Task descriptions and statistics of the ALUE benchmark. Test sets shown in bold use labels that have been made publicly available. The average sequence length, and standards deviation, are calculated based on the word count of the tokenized text of the training set.}
	\label{tab:alue_stat}
\end{table*}

Table~\ref{tab:pretrain_stat} shows the size of corpora before and after applying our data cleaning procedure.  Overall, our final pre-training corpus is 115GB of textual data, which represents 22\% of the original data. Despite the high ratio of discarded data, we show later that it will not have negative impact on the final model compared with prior works that performed light pre-processing~\cite{safaya2020kuisail,abdul2021arbert}. 

It is worth mentioning that our pre-training corpus has comparable size with the one of prior works like \asafaya{} and \marbert{} (95GB and 128GB respectively). Finally, we apply the Arabic text normalization procedure of \arabert\footnote{\url{https://github.com/aub-mind/arabert/blob/master/preprocess.py}} which includes removing emoji, tashkeel, tatweel, and html markup. We refer the readers to ~\cite{antoun2020arabert} for more details.

\subsection{Model and Implementation}

We use a byte-level byte pair encoding (BBPE)~\cite{wei2021training} tokenizer to process sub-tokens. BBPE first converts the text to a sequence of bytes and then builds BPE vocabulary~\cite{sennrich2016neural} on top of the byte-level representations. The authors show that BBPE eliminates the out-of-vocabulary problem and improves the learning of the representations of rare words. We set the vocabulary size to 64k, twice the one of \asafaya{} and \cmlbert{}, similar to \arabert{}, and 36\% less than \arbert{} and \marbert{}. 

\jaber{} and \saber{} has the same architecture and pre-training tasks as \bert{-base} and \bert{-large}~\cite{devlin2018bert} respectively. We pre-trained the models on both Masked Language Modelling (MLM) and Next Sentence Prediction (NSP) unsupervised tasks. In MLM, we use whole word masking with a probability of 15\%. The original tokens are replaced with the \textsc{[MASK]} special tokens with 80\% of the times, 10\% by a random token, while we keep the original token in the remaining 10\%. We used a duplication factor of 3 during data generation, meaning that each input sequence has 3 random sets of masked tokens. 

We perform pre-training on 16 servers\footnote{On Huawei Cloud \url{https://www.huaweicloud.com/product/modelarts}} for 15 and 5 epochs for \jaber{} and \saber{} respectively. Each server contains 8 NVIDIA Tesla V100 GPUs with 32GB of memory. The distributed training is achieved through Horovod ~\cite{sergeev2018horovod} with full precision. We set the initial learning rate to 1e-4, with 10000 warm-up steps, and used AdamW ~\cite{loshchilov2017decoupled} optimizer with a learning rate linear decay. We only train with the maximum sequence length of 128, while setting the per GPU batch size to 64 and 32 for \jaber{} and \saber{} respectively. It takes about 16 and 32 hours to finish one epoch for \jaber{} and \saber{} respectively.

\begin{table*}[!htp]
	
	\centering
	\resizebox{\textwidth}{!}{
		\begin{tabular}{l|ccccc|cc}
			\toprule
			& \bf \asafaya & \bf \arabert & \bf \cmlbert & \bf \arbert & \bf \marbert & \bf \jaber & \bf \saber \\
			
			\midrule
			
			\textbf{MQ2Q*} & 73.3$\pm$0.6 & 73.5$\pm$0.5 & 68.9$\pm$1.1 & 74.7$\pm$0.1  & 69.1$\pm$0.9 & \underline{75.1$\pm$0.3} & \bf 77.7$\pm$0.4\\
			
			\textbf{MDD} & 61.9$\pm$0.2 & 61.1$\pm$0.3 & 62.9$\pm$0.1 & 62.5$\pm$0.2  & 63.2$\pm$0.3  & \underline{65.7$\pm$0.3} & \bf 67.7$\pm$0.1  \\
			
			\textbf{SVREG} & 83.6$\pm$0.8 & 82.3$\pm$0.9 & 86.7$\pm$0.1 & 83.5$\pm$0.6  & \underline{88.0$\pm$0.4}  & 87.4$\pm$0.7 & \bf 89.3$\pm$0.3 \\
			
			\textbf{SEC}  & 42.4$\pm$0.4 & 42.2$\pm$0.6 & 45.4$\pm$0.5 & 43.9$\pm$0.6  &  \underline{47.6$\pm$0.9}  & 46.8$\pm$0.8 & \bf 49.0$\pm$0.5\\
			
			\textbf{FID}  & 83.9$\pm$0.6 & 85.2$\pm$0.2  & 84.9$\pm$0.6 & \underline{85.3$\pm$0.3}  & 84.7$\pm$0.4  & 84.8$\pm$0.3 & \bf 86.1$\pm$0.3\\
			
			\textbf{OOLD}  & 88.8$\pm$0.5  & 89.7$\pm$0.4  & 91.3$\pm$0.4 & 90.5$\pm$0.5  & 91.8$\pm$0.3  & \underline{92.2$\pm$0.5} & \bf 93.4$\pm$0.4\\
			
			\textbf{XNLI}  & 66.0$\pm$0.6 & 67.2$\pm$0.4  & 55.7$\pm$1.2 & 70.8$\pm$0.5  & 63.3$\pm$0.7 & \underline{72.4$\pm$0.7} & \bf 75.9$\pm$0.3 \\
			
			\textbf{OHSD}  & 79.3$\pm$1.0  & 79.9$\pm$1.8 & 81.1$\pm$0.7 & 81.9$\pm$2.0  & 83.8$\pm$1.4  & \underline{85.0$\pm$1.6} & \bf 88.9$\pm$0.3 \\
			
			\midrule
			\textbf{Avg.}  & 72.4$\pm$0.6 & 72.6$\pm$0.6 & 72.1$\pm$0.6 & 74.1$\pm$0.6 & 73.9$\pm$0.7  & \underline{76.2$\pm$0.7} & \bf 78.5$\pm$0.3\\
			
			\bottomrule
			
		\end{tabular}
	}
	\caption{\textsc{Dev} performances and standard deviations over 5 runs on the ALUE benchmark. Bold entries describe the best results among all models, while underlined entries show best results among \bert{-base} models. * indicates that the results are on our own MQ2Q dev set.}
	\label{tab:dev_alue}
\end{table*}

\section{Experiments}
\label{sec:Experiments}

\subsection{Datasets}

We run experiments on eight tasks from
the ALUE benchmark~\cite{seelawi2021alue}. It is a newly proposed benchmark that gathers a diversified collection of Arabic NLU tasks: 4 single-sentence, 2 sentence-pair, and one multi-label classification tasks, as well as a single regression task. The final score is the unweighted average over the eight tasks. We refer the readers to~\cite{seelawi2021alue} for detailed descriptions of ALUE datasets.

As Table~\ref{tab:alue_stat} shows, 5 (out of 8) ALUE tasks are sourced from Tweets, and 6 tasks contains Arabic dialect data. This makes ALUE a suitable tool to identify useful models and keep track of the progress in the Arabic NLU field. However, ALUE training datasets and their sentence lengths are relatively small compared to English GLUE~\cite{wang2018glue}. In addition, three tasks (FID, MQ2Q, XNLI) are not supported by a dev set, and the test set labels are publicly provided for three tasks (MDD, FID, XNLI). 

We use a simple yet generic method to obtain a dev set for the MQ2Q task\footnote{Following ALUE paper, we treat FID and XNLI test set as a dev set.}. First, we translated the development set of QQP task\footnote{\url{https://www.quora.com/q/quoradata/First-Quora-Dataset-Release-Question}} from English to Arabic using an online translation service. Then we randomly selected 2k positive and negative samples (4k in total). In order to ensure a high-quality corpus, we only select sentence pairs that don't contain English alphabet letters. This set is inclusively used as a proxy to evaluate models and select the best one for test submission.

Furthermore, we also consider \aner{}~\cite{benajiba2007anersys} for evaluation. It is a well-established benchmark for Arabic Named Entity Recognition (NER) which includes 4 types of named-entities. We run experiments on the train/test split provided by ~\cite{obeid2020camel} and report mention-level F1 scores using the official \conll{-2003}~\cite{tjong2003introduction} evaluation script\footnote{\url{https://www.clips.uantwerpen.be/conll2000/chunking/conlleval.txt}}.

\subsection{Finetuing Details}

We run extensive experiments in order to fairly compare \jaber{}\footnote{as well as for fine-tuning \saber{}} with \asafaya{}, \arabert{}, \cmlbert{}, \arbert{} and \marbert{} on the ALUE tasks. For all these models, we use AdamW optimizer with learning rate with linear decay. We search\footnote{We used grid search with multiple runs} the learning rate from \{7e-6, 2e-5, 5e-5\}, batch size from \{8, 16, 32, 64, 128\}, hidden dropout from \{0.1, 0.2, 0.3, 0.4\}, and fixed the epoch number to 30. The aforementioned HP search strategy is applied to all models, and the best hyper-parameters are listed in Table~\ref{tab:hp_cross} in Appendix~\ref{sec:ALUE Hyper-parameters}. 

In order to validate the statistical significance of our results, we run all experiments 5 times with different random seeds, and we report average scores and standards deviations. For \jaber{} and \saber{} test submissions, we use the models performing the best on the dev set for each task. Our fine-tuning code is based on the PyTorch~\cite{paszke2019pytorch} version of the HuggingFace Transformers~\cite{wolf2020transformers} library. We run all experiments on a single NVIDIA Tesla V100 GPU. 

\begin{table*}[!htp]
	
	\centering
	\begin{tabular}{l|cccccccc|c}
		\toprule
		\textbf{Model} & \textbf{MQ2Q}& \textbf{MDD}& \textbf{SVREG}& \textbf{SEC}& \textbf{FID}& \textbf{OOLD}& \textbf{XNLI}& \textbf{OHSD}& \textbf{Avg.}\\
		\midrule
		
		m\bert{} & 83.2 & 61.3 & 33.9 & 14.0 & 81.6 & 80.3 & 63.1 & 70.5 & 61.0 \\
		\asafaya & 85.7 & 59.7 & 55.1 & 25.1 & 82.2 & 89.5 & 61.0 & 78.7 &  67.1 \\
		
		\midrule
		\jaber & 93.1 & 64.1 & 70.9 & 31.7 & 85.3 & 91.4 & 73.4 & 79.6 & 73.7 \\ 
		
		\saber & \bf 93.3 & \bf 66.5 & \bf 79.2 & \bf 38.8 & \bf 86.5 & \bf 93.4 & \bf 76.3 & \bf 84.1 & \bf 77.3 \\ 
		\bottomrule
		
	\end{tabular}
	\caption{Leaderboard test results (as of 03/01/2022) of experiments on ALUE tasks. Bold entries show the best results among all models.}
	\label{tab:test_alue}
\end{table*}

\subsection{Results}

Table~\ref{tab:dev_alue} shows the dev set performance of models trained on ALUE tasks. For each model, we report the average and standard deviation of 5 runs. First, we notice that variance in performances of multiple runs is roughly the same on average for all \bert{-base} models. The variance is within an acceptable range (0.6-0.7 on average), except on OHSD where all models suffer from high variance. On the other hand, the \bert{-large} \saber{} model has a significantly lower variance (about half) on all the tasks compared to \bert{-base} models.  

Second, we notice that \asafaya{} and \arabert{} perform roughly the same with 72.4\% and 72.5\% on average respectively. This might be because both models have similar training data sizes. \asafaya{} had 95GB of text data that were duplicated 3 times (285GB), while \arabert{} had 27GB duplicated 10 times (270GB). Third, we observe that \marbert{} performs well on Tweets tasks, but less so on MSA tasks and vice versa for \arbert{}. 

Although \cmlbert{} significantly outperforms \asafaya{}, \arabert{}, and \arbert{} on 6, 5, and 4 tasks respectively, its overall performance is not competitive to the other baselines (72.1\% on average). This overall lower performance can be attributed to the lower performance of \cmlbert{} on MQ2Q (68.9) and XNLI (55.7), which are sentence pair classification tasks and consists of MSA data.

\jaber{} significantly outperforms \arbert{} and \marbert{} by 2.1\% and 2.3\% on overall average ALUE score respectively. \marbert{} reported a  higher score than \jaber{} on SVREG (88.0\% vs. 87.4\%) and SEC (47.6\% vs. 46.8\%). However, \jaber{} significantly outperforms this particular model on MSA tasks by +9.1\% and +6.0\% on XNLI and MQ2Q respectively. Furthermore, it shows better performances on the remaining dialect and tweet based tasks. Expectedly, \saber{} significantly outperforms \jaber{} by a margin of 3.6\% on average, as well as all other \bert{-base} models on all the ALUE tasks.

The results are promising, especially when we consider that our pre-training data did not contain tweets data, and we pre-trained our model with fewer data and fewer epochs compared to \marbert{}. Moreover, the fact that a single model (\jaber{}) works well on MSA, dialect, and tweets tasks is an indicator that our models have potential to generalize well independently from the source data.

Table~\ref{tab:test_alue} shows the performances of the top 4 models submitted to ALUE leaderboard\footnote{\url{https://www.alue.org/leaderboard}} by 03/01/2022. \jaber{} outperforms \asafaya{}\footnote{Submitted by the authors of the ALUE paper.} by 6.6\% on average compared with 3.2\% on the dev set. \jaber{} astonishingly outperforms \asafaya{} on SVREG, XNLI, MQ2Q and SEC by 15.8\%, 12.4\%, 7.5\% and 6.6\% respectively. This may be because the private sample sets were collected at different time frames from train and dev set ~\cite{seelawi2021alue}, and also are designed to be harder. 

Similar to the dev set scores, \saber{} significantly outperforms \jaber{} by 3.6\% on the average test results as well, therefore \saber{} reports state-of-the-art results on ALUE. Unfortunately, we could not submit the remaining baselines to the leaderboard due to the rules\footnote{\url{https://www.alue.org/FAQ}} defined by the ALUE toolkit owners. 

\begin{table}[!ht]
	
	\begin{center}
		
		\begin{tabular}{l|l}
			\toprule
			
			Model & F1 score  \\
			
			\midrule
			\marbert & 80.50$\pm$0.35\\
			\asafaya & 82.05$\pm$0.28\\ 
			\cmlbert & 82.53$\pm$0.21\\
			\arabert & 82.72$\pm$0.23\\
			\arbert & 84.03$\pm$0.22\\
			\midrule
			\jaber & \bf 84.20$\pm$0.32\\
			\bottomrule
		\end{tabular}
		
	\end{center}	
	\caption{Test set mention Level F1 scores of Arabic \bert{-base} models fine-tuned on \aner. }
	\label{tab:res_ner}
\end{table}

To further validate our approach, we perform an evaluation on a sequential labeling task, namely named entity recognition (NER). Table~\ref{tab:res_ner} shows models F1 mention level score on the test set of \aner{} corpus over 5 runs. Consistent with the results obtained on ALUE, \jaber{} reports the highest score of 84.2\% and outperforms all its counterpart \bert{-base} models, while their standard deviation indicate that the improvement is significant. Expectedly, \marbert{} is the worst model on this task (80.5\%) because the data was sourced from MSA news articles.

\section{Conclusion and Future Work}
\label{sec:Conclusion}

In this work, we provide detailed information of the steps we followed to pre-train 2 new Arabic BERT models. We performed a systematic evaluation with previously existing models in the field. Our experiment shows that \jaber{} significantly outperforms several baselines which are pre-trained under similar settings. Also, \saber{} sets a new state-of-the-art on the ALUE benchmark, a collection of 8 diversified Arabic NLU tasks.

In future, we will work on enhancing the \textit{dialect awareness} of our models by pre-training it on a massive amount of Tweets data as done by \marbert{}~\cite{abdul2021arbert}. Also, we would like to explore more pre-training architectures and task formulations like T5~\cite{raffel2019exploring} and GPT-3~\cite{brown2020language} for Arabic NLU. We make the source code and pre-trained weights of \jaber{} freely available at
\href{https://github.com/huawei-noah/Pretrained-Language-Model/tree/master/JABER-PyTorch}{https://github.com/huawei-noah/Pretrained-Language-Model/JABER-PyTorch}.

\section*{Acknowledgments}
We thank Mindspore\footnote{\url{https://www.mindspore.cn/}}, a new deep learning computing framework, for the partial support of this work.

\normalem

\bibliography{custom}
\bibliographystyle{acl_natbib}

\clearpage
\newpage
\appendix

\section{Filtering Heuristics}
\label{sec:Preprocessing}

\begin{enumerate}
	\item Remove sentences with HTML or Javascript code~\cite{raffel2019exploring}. 
	\item Remove sentences if it has less than 70\% Arabic characters. 
	\item Remove sentences with less than 8 words.
	\item Remove sentences with more than 3 successive punctuation (excluding dot).
	\item Remove document with less than 64 words.
	\item Remove long spans of non-Arabic text (mostly English) inside a sentence. We observe that most of these sentences where garbage text and not related with the content.
	\item Represent each sentence by the concatenation of the first and last 3 words\footnote{We considered only  words that do not include digits and has more than 3 characters.}. We de-duplicate the corpus by only keeping the first occurrence of sentences with the same key.
	\item Discard a document if more than 30\% of its sentences are discarded by the last step.
\end{enumerate}

\section{ALUE Hyper-parameters}
\label{sec:ALUE Hyper-parameters}

\begin{table*}[!th]
	\centering
	\begin{tabular}{l|c c c c c c c c }
		
		\toprule
		\textbf{Model} & \textbf{MQ2Q}& \textbf{MDD}& \textbf{SVREG}& \textbf{SEC}& \textbf{FID}& \textbf{OOLD}& \textbf{XNLI}& \textbf{OHSD}\\
		
		\midrule
		\multicolumn{9}{c}{\textit{\asafaya}}\\
		\midrule
		batch size & 64 & 16 & 16 & 16 & 32 & 32 & 64 & 16\\
		hidden dropout  & 0.1 & 0.1 & 0.1 & 0.1 & 0.1 & 0.1 & 0.1 & 0.1\\
		learning rate  & 2e-05 & 2e-05 & 2e-05 & 2e-05 & 2e-05 & 2e-05 & 2e-05 & 2e-05\\
		
		\midrule
		\multicolumn{9}{c}{\textit{\arabert}}\\
		\midrule
		batch size & 128 & 32 & 8 & 8 & 8 & 32 & 32 & 16 \\
		hidden dropout  & 0.1 & 0.1 & 0.2 & 0.1 & 0.1 & 0.1 & 0.3 & 0.1\\
		learning rate  & 2e-05 & 2e-05 & 2e-05 & 2e-05 & 2e-05 & 2e-05 & 2e-05 & 2e-05\\
		
		\midrule
		\multicolumn{9}{c}{\textit{\cmlbert}}\\
		\midrule
		batch size & 16 & 8 & 8 & 32 & 8 & 128 & 32 & 8 \\
		hidden dropout  & 0.2 & 0.2 & 0.2 & 0.1 & 0.2 & 0.1 & 0.1 & 0.1\\
		learning rate  & 5e-05 & 2e-05 & 2e-05 & 5e-05 & 2e-05 & 2e-05 & 2e-05 & 2e-05\\
		
		\midrule
		\multicolumn{9}{c}{\textit{\arbert}}\\
		\midrule
		batch size & 64 & 16 & 32 & 8 & 32 & 128 & 32 & 32\\
		hidden dropout & 0.1 & 0.1 & 0.3 & 0.3 & 0.1 & 0.1 &  0.1 & 0.3\\
		learning rate & 2e-05 & 2e-05 & 2e-05 & 2e-05 & 2e-05 & 2e-05 & 2e-05 & 7e-06\\
		
		\midrule
		\multicolumn{9}{c}{\textit{\marbert}}\\
		\midrule
		batch size & 64 & 64 & 16 & 8 & 64 & 64 & 64 & 64 \\
		hidden dropout  & 0.3 & 0.2 & 0.1 & 0.3 & 0.1 & 0.2 & 0.2 & 0.1\\
		learning rate  & 2e-05 & 2e-05 & 2e-05 & 2e-05 & 2e-05 & 2e-05 & 2e-05 & 2e-05\\
		
		\midrule
		\multicolumn{9}{c}{\textit{\jaber}}\\
		\midrule
		batch size & 64 & 32 &  8 & 16 & 32 & 128 & 16 & 32\\
		hidden dropout & 0.3  & 0.2 & 0.1 & 0.1 & 0.1 & 0.2 & 0.1 & 0.3 \\
		learning rate & 2e-05 & 2e-05 & 2e-05 & 2e-05 & 2e-05 & 2e-05 & 2e-05 & 7e-06\\
		
		\midrule
		\multicolumn{9}{c}{\textit{\saber}}\\
		\midrule
		batch size & 32 & 32 &  8 & 8 & 32 & 32 & 32 & 32\\
		hidden dropout & 0.1  & 0.1 & 0.2 & 0.2 & 0.3 & 0.2 & 0.2 & 0.1  \\
		learning rate & 7e-06 & 2e-05 & 7e-06 & 2e-05 & 2e-05 & 7e-06 & 7e-06 & 7e-06\\
		
		\bottomrule
	\end{tabular}
	\caption{For each ALUE task, the value of best Hyperparameters for Arabic \bert{} models.}
	\label{tab:hp_cross}
\end{table*}

\end{document}